\pdfoutput=1

\documentclass[11pt]{article}

\usepackage[review]{ACL2023}

\usepackage{times}
\usepackage{latexsym}

\usepackage[T1]{fontenc}

\usepackage[utf8]{inputenc}

\usepackage{microtype}

\usepackage{inconsolata}

\title{DuanzAI: Slang-Enhanced LLM with Prompt for Humor Understanding }

\author{    
    Yesian Rohn\textsuperscript{1}
    \\
    \\
    \textsuperscript{1}School of Computer Science, Fudan University \\}

\usepackage{graphicx}

\begin{document}
\maketitle
\begin{abstract}
Language's complexity is evident in the rich tapestry of slang expressions, often laden with humor and cultural nuances. This linguistic phenomenon has become increasingly prevalent, especially in digital communication. However, existing AI models, including ChatGPT-3.5, face challenges in comprehending these nuances, particularly in Chinese slang. In this study, we present DuanzAI, an innovative approach enhancing Large Language Models (LLMs) with deep Chinese slang comprehension. Leveraging curated datasets and advanced techniques, DuanzAI bridges the gap between human expression and AI comprehension, enabling contextually relevant responses. Our experiments contrast LLMs' performance with a custom Punchline Entity Recognition (PER) system, integrating phonetic matching and pinyin2hanzi techniques. Applying these insights, we developed ChatDAI, an advanced chatbot and released our code at \url{https://github.com/YesianRohn/DuanzAI}.
\end{abstract}

\section{Introduction}  

\begin{quote}
\textit{"I think the next best thing to solving a problem is finding some humor in it."}
  
\raggedleft 
{- \citet{Frank01}}
\end{quote}

Language, as a dynamic and evolving entity, manifests itself in various forms within different social contexts. One of the fascinating linguistic phenomena that has emerged prominently in recent years is slang. Slang, defined as an informal language form primarily used within specific social groups, encompasses words or phrases that are metaphorical, pun-based, or even contrary to their literal meanings. It thrives on metaphor, wordplay, and implicit implications, adding depth and nuance to communication. Traditionally confined to oral interactions, the digital era has witnessed a surge in slang usage, especially on platforms like instant messaging applications and social media networks. This proliferation has sparked a pressing need for artificial intelligence systems to comprehend and respond effectively to these informal linguistic expressions.

Understanding slang is not merely a matter of linguistic curiosity; it holds substantial importance in the realm of natural language processing, particularly in the development of conversational AI. Humor, emotions, and subtle social cues often rely on slang usage, making it imperative for AI systems to grasp these nuances to facilitate more authentic and engaging interactions.

\begin{figure}
\centering 
\includegraphics[width=0.5\textwidth]{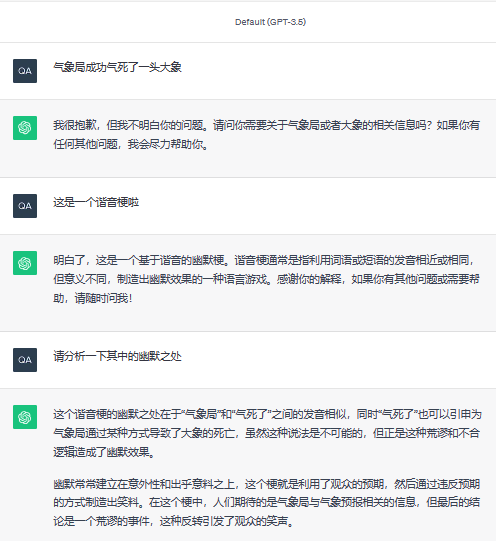}
\caption{ChatGPT-3.5 cannot understand Chinese homophonic pun jokes.} 
\label{fig:figure1} 
\end{figure}

Despite its prevalence, the study of slang in the context of natural language processing remains relatively unexplored. Existing research efforts have begun to delve into the applications of slang, but the field still lacks comprehensive attention. Slang's inherent ambiguity, metaphorical nature, and irregularities pose significant challenges for artificial intelligence systems. Current conversational AI, including state-of-the-art models like GPT-3.5, often struggle to comprehend and respond appropriately to Chinese slang expressions, leading to a gap in humor understanding and cultural relevance (Figure ~\ref{fig:figure1}).

Therefore, we create DuanzAI, an innovative approach aimed at enhancing LLMs with a deep understanding of Chinese slang. Leveraging a carefully curated dataset of Chinese slang expressions and optimizing existing pre-trained language models, DuanzAI seeks to bridge the gap between human expression and AI comprehension. By incorporating slang-specific prompts, DuanzAI aims to enable AI systems not only to recognize and interpret diverse slang expressions but also to respond contextually, thereby elevating the AI's ability to understand humor, emotions, and social dynamics within conversations.

When we attempt to engage LLM in understanding humor, their cognitive process tends to revolve around identifying punchlines, highlighting the contrasts within these punchlines, and subsequently conducting analysis. In the first and third steps, they demonstrate relatively satisfactory performance on their own; however, errors in the second step can also impact the outcome of the third step. Moreover, if both the first and second steps are accurate, leveraging the extensive knowledge base of large models would generally ensure reliable analysis.

Consequently, we initiated an experiment to assess the performance of LLMs, like ChatGLM-6B\cite{zeng2022glm} and ChatGPT3.5 in identifying punchlines. This was then contrasted with our self-constructed PER system. Following this, we proceeded with the second step by utilizing a combination of online phonetic matching API and the pinyin2hanzi\cite{pinyin2hanzi} technique to locate the original words. Subsequently, we integrated PER and the aforementioned components into prompts to observe the enhanced performance of the large model. Based on the research on humor comprehension mentioned above, we constructed our ChatDAI: DuanzAI Chatbot using the Spark\cite{spark} large model API.

\section{Datasets and Task Setups}

\subsection{Datasets}

Our research leverages a substantial dataset of Chinese homophonic puns and slang expressions obtained from the rich linguistic landscape of Sina Weibo, a prominent social media platform in China. This dataset comprises nearly 2500 meticulously curated instances, each meticulously annotated for both the punchline (the humorous element) and the original word or phrase. By focusing on these specific aspects, we aim to capture the essence of the humor embedded within slang expressions, enabling a detailed analysis of linguistic wit and creativity.

\subsection{Task Setups}

Our experimental design is meticulously crafted, encompassing three distinct phases to dissect the intricacies of humor recognition within Chinese slang expressions. These phases were structured as follows:

\subsubsection{Punchline Entity Recognition}
In the initial phase, we focus on developing a robust PER system. This specialized system was engineered to identify and extract the precise entities constituting the punchlines within the Chinese slang expressions. Leveraging a combination of linguistic analysis and contextual understanding, PER aimed to capture the essence of humor by pinpointing the specific words or phrases that constituted the punchlines. Through meticulous annotation and validation, we curated a dataset that served as the gold standard for punchline identification, forming the basis for our subsequent evaluations.

\subsubsection{Punchline Original Word Retrieval}
Building upon the identified punchlines, the second phase of our experiment involved the intricate task of locating the original words or phrases corresponding to these punchlines. To achieve this, we employed a sophisticated combination of online phonetic matching APIs and the pinyin2hanzi technique. Phonetic matching was utilized to generate candidate words based on their pronunciation, which were then rigorously validated using pinyin2hanzi, a method converting phonetic representations (pinyin) back into accurate Chinese characters (hanzi). This meticulous process ensured the accurate retrieval of the original words associated with the identified punchlines, establishing a reliable ground truth for our evaluations.

\subsubsection{Comprehensive Humor Understanding}
In the final phase of our experiment, we orchestrated a comprehensive evaluation of humor understanding within the context of Chinese slang expressions. This phase involved integrating the insights derived from the PER system and the precise original word retrieval into specially crafted prompts. These prompts were designed to challenge pre-trained language models, including ChatGLM-6B and ChatGPT3.5, to comprehend the nuanced layers of humor encapsulated within the slang expressions. By infusing the prompts with the knowledge garnered from PER and the phonetic matching techniques, our objective was to discern subtle distinctions in humor recognition between the pre-trained models and our specialized system.

Through these meticulously designed phases, we sought to unravel the subtle differentiations in punchline recognition between pre-trained models and our expertly crafted systems. This multifaceted approach provids a robust framework for evaluating the humor comprehension capabilities of both traditional language models and our specialized systems, shedding light on the complexities of humor recognition in the realm of Chinese slang expressions.

\section{Experiments}

\subsection{Punchline Entity Recognition}

In the initial phase of our experiments, we aim to assess the punchline recognition abilities of LLMs. Additionally, we sought to address the challenge of punchline entity recognition through a direct and simple approach, utilizing BERT-LSTM-CRF (Bidirectional Encoder Representations from Transformers - Long Short-Term Memory - Conditional Random Fields). This methodology, derived from named entity recognition (NER) techniques \cite{9039685}, provided a robust framework for solving the punchline entity recognition problem. To evaluate the capabilities of ChatGPT and ChatGLM in recognizing punchlines within Chinese slang expressions, we employed a assessment with Exact Match Accuracy(EMA) and Similar Match Accuracy(SMA).  Concurrently, we explored the performance of our specialized PER system based on BERT-LSTM-CRF.  The evaluation metrics included precision, recall, and F1-score, offering a comprehensive understanding of each model's accuracy in punchline recognition.

The results of our experiments, presented in Table ~\ref{tab:accents}, which is evident from the table that our PER system, utilizing the BERT-LSTM-CRF framework, outperforms both ChatGPT and ChatGLM in punchline entity recognition. The higher EMA achieved by our specialized system demonstrates its superior accuracy and robustness in capturing punchlines within Chinese slang expressions.

\begin{table}
\centering
\begin{tabular}{lcc}
\hline
\textbf{Model} & \textbf{EMA} & \textbf{SMA} \\
\hline
\verb|ChatGLM-6B (0-Shot)| & 0.57 & 0.73 \\
\verb|GPT3.5 (0-Shot)| & 0.87 & 0.95 \\
\verb|GPT3.5 (5-Shot)| & 0.92 & 0.97 \\
\verb|GPT4 (5-Shot)| & 0.92 & 0.97 \\
\verb|Ours| & 0.97 & 0.98 \\
\hline
\end{tabular}
\caption{Results of Punchline Entity Recognition in All Models Experiments.}
\label{tab:accents}
\end{table}

\begin{table}
\centering
\begin{tabular}{lcc}
\hline
\textbf{Precision} & \textbf{Recall} & \textbf{F1-Score} \\
\hline
\verb|0.970|     &\verb|0.969| &\verb|0.969|  \\

\hline
\end{tabular}
\caption{Results of BERT-LSTM-CRF Punchline Entity Recognition Experiments.}
\label{tab:ourmodel}
\end{table}

\subsection{Comprehensive Humor Understanding}

In the pursuit of enhancing humor comprehension within Chinese slang expressions, we adopted two distinct approaches: Clue Provided and 5-Shot. These methodologies were designed to augment the capabilities of our language models, specifically ChatGPT 3.5, in grasping the nuances of humor. To evaluate the effectiveness of these approaches, the output generated by our enhanced models was subjected to human evaluation. Participants were asked to assign scores based on the perceived humor understanding, with a maximum score of 100 indicating a perfect understanding of the humor embedded in the expressions.

\subsubsection{Clue Provided Approach}

In the Clue Provided scenario, the language model was furnished with additional context or clues related to the punchline. This approach aimed to simulate a scenario where subtle hints or off-site information was available, aiding the model in deciphering the humor within the slang expressions. The Clue Provided method resulted in an enhanced humor comprehension, as evidenced by the model's improved score to 53.1 out of 100, compared to the baseline of 39.3 achieved by ChatGPT 3.5 in the 0-Shot setting.

\subsubsection{5-Shot Approach}

In the 5-Shot approach, the language model was provided with five iterations or examples of similar slang expressions to establish a pattern. This technique leveraged a small dataset to facilitate learning and improve the model's grasp of the underlying humor structure. Despite the limited training data, the 5-Shot method yielded promising results, with ChatGPT 3.5 achieving a humor comprehension score of 51 out of 100.

\subsubsection{Human Evaluation Results}

The results of the human evaluation are summarized in Table ~\ref{tab:hf}. The scores obtained in the Clue Provided and 5-Shot approaches highlight the significant improvement in humor understanding compared to the 0-Shot scenario. These evaluations underscore the effectiveness of providing additional context or limited examples to enhance the language model's ability to comprehend humor within Chinese slang expressions.

\begin{table}
\centering
\begin{tabular}{lc}
\hline
\textbf{Approach} & \textbf{Score (100)} \\
\hline
\verb|GPT3.5(0-Shot)| & 39.3 \\
\verb|GPT3.5(clue-provided)| & 53.1 \\
\verb|GPT3.5(5-Shot)| & 51.0 \\

\hline
\end{tabular}
\caption{Human Evaluation Scores for Comprehensive Humor Understanding.}
\label{tab:hf}
\end{table}

These findings demonstrate the positive impact of contextual clues and limited training examples on the language model's humor comprehension abilities. The Clue Provided and 5-Shot approaches offer valuable insights into how contextual information and minimal training data can substantially enhance the nuanced understanding of humor within the realm of Chinese slang expressions, paving the way for more culturally sensitive and engaging AI interactions.

\section{Related Work}

Slang detection involves the identification of informal language expressions within a given text, utilizing natural language processing techniques and machine learning algorithms to discern the differences between slang and formal expressions \cite{pei2019slang}. Simple and practical slang detection methods include emotion analysis techniques based on constructing dictionaries \cite{dhuliawala2016slangnet,wu2018slangsd} and modular multi-factor collection \cite{pal2015detection,gupta2019slangzy}. However, these approaches heavily rely on static information and structure of slang expressions, lacking the ability to comprehend the flexible semantics of slang and handle newly emerging expressions.

More sophisticated research in slang detection revolves around deep learning-based automatic slang detection and cognitive usage frameworks. Researchers have leveraged techniques such as multi-layer bidirectional LSTM perceptrons (MLP) \cite{kulkarni2018simple}, conditional random fields (CRF) \cite{lafferty2001conditional}, and Part-of-Speech (POS) taggers \cite{korolainen2014part} to extract comprehensive linguistic features of slang. The field has seen continuous improvements, including:

\begin{enumerate}
    \item \textbf{Sequence Models:} Researchers have explored dialectal variations for feature extraction, leading to sequence models like the LID method, enabling recognition of diverse dialects \cite{jurgens2017incorporating,lourentzou2019adapting}. Dialectal varieties, being a significant source of slang, are well recognized by sequence models \cite{kulkarni2018simple}.
    
    \item \textbf{Character-level Convolution:} Unlike traditional methods based on word embeddings, character-level convolution is effective in capturing correlated words and morphological variations in text \cite{zhang2015character,bu2018hybrid}.
    
    \item \textbf{Pre-Training Models:} In specific subtasks like humor detection, recent studies have incorporated pre-training models such as BERT, fine-tuning them on training data for tasks like joke scoring \cite{mao2019bert,annamoradnejad2020colbert}.
\end{enumerate}

In the current era of large-scale models, it appears that fine-tuning existing models with carefully crafted prompts can lead to more successful outcomes in achieving the ability to comprehend humor. The best paper presented at ACL 2023 assessed the level of humor "understanding" in large models through meticulously designed prompts. This contribution showcases the effectiveness of prompt engineering in evaluating the humor comprehension capabilities of these models 
\cite{hessel-etal-2023-androids}.

\section{Application}

Building upon the insights gained from our research on humor comprehension, we embarked on the practical application of our findings. The culmination of our efforts led to the creation of ChatDAI: DuanzAI Chatbot, a sophisticated conversational agent designed to understand and engage in humorous interactions within the realm of Chinese slang expressions.

\subsection{Development Process}
Our chatbot development process was informed by the research findings mentioned earlier. We began by contrasting the capabilities of ChatGPT and our PER system. Following this comparative analysis, we seamlessly integrated the PER system with a combination of online phonetic matching APIs and the pinyin2hanzi technique. This integration allowed us to accurately locate the original words corresponding to punchlines, providing a robust foundation for humor comprehension.

\subsection{Integration of Components}
The culmination of our research efforts resulted in the integration of PER, phonetic matching, and pinyin2hanzi components into prompts used by our chatbot. By incorporating these elements, ChatDAI was equipped with an enhanced understanding of humor, allowing it to decipher the complexities of Chinese slang expressions. This integration significantly amplified the chatbot's ability to engage users in witty and culturally relevant conversations.

\subsection{ChatDAI Implementation}
Utilizing the Spark large model API, we brought our vision to life with the creation of ChatDAI. It serves as an embodiment of our research, showcasing the practical implications of our work. ChatDAI's ability to comprehend humor within Chinese slang expressions is a testament to the advancements made in natural language processing, humor recognition, and conversational AI.

\subsection{Limitations and Choice of Framework}
It's noteworthy that we chose Spark as the baseline framework for ChatDAI, even though ChatGPT might offer superior performance. This decision was primarily influenced by OpenAI API limitations and regulatory constraints within the context of China. These restrictions necessitated the utilization of Spark as a viable and practical alternative, allowing us to implement our vision while adhering to local regulations.

\subsection{ChatDAI in Action}
You can experience the capabilities of ChatDAI by visiting our live chatbot at www.duanzai.top. The chatbot's interface provides users with an opportunity to engage in humorous conversations, exploring the richness of Chinese slang expressions in an interactive and engaging manner.

\subsection{Humor Comprehension Visualization}
\begin{figure}
\centering 
\includegraphics[width=0.5\textwidth]{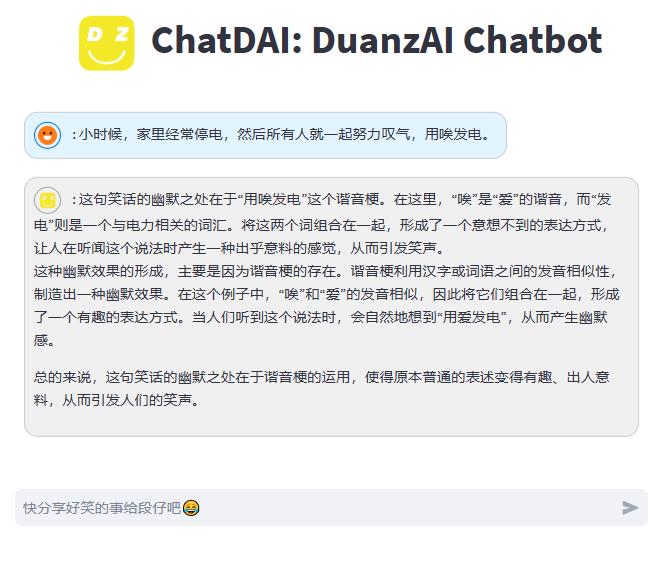}
\caption{Chat with ChatDAI.} 
\label{fig:figure2} 
\end{figure}

Figure ~\ref{fig:figure2} provides a visual representation of ChatDAI's humor comprehension abilities. The graph showcases the chatbot's proficiency in understanding and responding to various slang expressions, highlighting its capacity to navigate the intricacies of humor within the Chinese language.

\section{Conclusion}

In this paper, we pointed out that current large language models exhibit room for improvement in humor comprehension, addressed the understanding problem of Chinese humor and enhanced the humor comprehension abilities of large language models through the incorporation of carefully crafted prompt pipelines. The approach presented in this study establishes a solid baseline for future research endeavors aimed at refining and advancing the capabilities of language models in understanding humor, particularly in the context of Chinese linguistic expressions.

\bibliography{anthology,custom}

\begin{thebibliography}{20}
\expandafter\ifx\csname natexlab\endcsname\relax\def\natexlab#1{#1}\fi

\bibitem[{Annamoradnejad and Zoghi(2020)}]{annamoradnejad2020colbert}
Iman Annamoradnejad and Ghazaleh Zoghi. 2020.
\newblock Colbert: Using bert sentence embedding for humor detection.
\newblock \emph{arXiv preprint arXiv:2004.12765}, 1(3).

\bibitem[{Bu and Cho(2018)}]{bu2018hybrid}
Shi~Jie Bu and Sang-Bum Cho. 2018.
\newblock A hybrid deep learning system of cnn and lrcn to detect cyberbullying from sns comments.
\newblock In \emph{Hybrid Artificial Intelligent Systems}, pages 561--572. Springer.

\bibitem[{Clark()}]{Frank01}
Frank~A. Clark.
\newblock I think the next best thing to solving a problem is finding some humor in it.

\bibitem[{Dhuliawala et~al.(2016)Dhuliawala, Kanojia, and Bhattacharyya}]{dhuliawala2016slangnet}
Shefali Dhuliawala, Debanjan Kanojia, and Pushpak Bhattacharyya. 2016.
\newblock Slangnet: A wordnet like resource for english slang.
\newblock In \emph{Proceedings of the Tenth International Conference on Language Resources and Evaluation (LREC'16)}, pages 4329--4332.

\bibitem[{Gupta et~al.(2019)Gupta, Taneja, Malik, Vij, Tayal, and Jain}]{gupta2019slangzy}
Akshay Gupta, Sanya~B Taneja, Gaurav Malik, Shruti Vij, Dinesh~K Tayal, and Anjali Jain. 2019.
\newblock Slangzy: A fuzzy logic-based algorithm for english slang meaning selection.
\newblock \emph{Progress in Artificial Intelligence}, 8:111--121.

\bibitem[{Hessel et~al.(2023)Hessel, Marasovic, Hwang, Lee, Da, Zellers, Mankoff, and Choi}]{hessel-etal-2023-androids}
Jack Hessel, Ana Marasovic, Jena~D. Hwang, Lillian Lee, Jeff Da, Rowan Zellers, Robert Mankoff, and Yejin Choi. 2023.
\newblock \href {https://doi.org/10.18653/v1/2023.acl-long.41} {Do androids laugh at electric sheep? humor {``}understanding{''} benchmarks from the new yorker caption contest}.
\newblock In \emph{Proceedings of the 61st Annual Meeting of the Association for Computational Linguistics (Volume 1: Long Papers)}, pages 688--714, Toronto, Canada. Association for Computational Linguistics.

\bibitem[{iFLYTEK(2023)}]{spark}
iFLYTEK. 2023.
\newblock spark-ai-python.
\newblock \url{https://github.com/iflytek/spark-ai-python}.

\bibitem[{Jurgens et~al.(2017)Jurgens, Tsvetkov, and Jurafsky}]{jurgens2017incorporating}
David Jurgens, Yulia Tsvetkov, and Dan Jurafsky. 2017.
\newblock Incorporating dialectal variability for socially equitable language identification.
\newblock In \emph{Proceedings of the 55th Annual Meeting of the Association for Computational Linguistics (Volume 2: Short Papers)}, pages 51--57.

\bibitem[{Korolainen(2014)}]{korolainen2014part}
Ville Korolainen. 2014.
\newblock Part-of-speech tagging in written slang.
\newblock \emph{Journal of Linguistics}, 50(2):343--371.

\bibitem[{Kulkarni and Wang(2018)}]{kulkarni2018simple}
Vivek Kulkarni and William~Yang Wang. 2018.
\newblock Simple models for word formation in slang.
\newblock In \emph{Proceedings of the 2018 Conference of the North American Chapter of the Association for Computational Linguistics: Human Language Technologies, Volume 1 (Long Papers)}, pages 1424--1434.

\bibitem[{Lafferty et~al.(2001)Lafferty, McCallum, and Pereira}]{lafferty2001conditional}
John Lafferty, Andrew McCallum, and Fernando~C Pereira. 2001.
\newblock Conditional random fields: Probabilistic models for segmenting and labeling sequence data.
\newblock \emph{Machine Learning}, 46(1-3):389--403.

\bibitem[{letiantian(2015)}]{pinyin2hanzi}
letiantian. 2015.
\newblock Pinyin2hanzi.
\newblock \url{https://github.com/letiantian/Pinyin2Hanzi}.

\bibitem[{Li et~al.(2022)Li, Sun, Han, and Li}]{9039685}
Jing Li, Aixin Sun, Jianglei Han, and Chenliang Li. 2022.
\newblock \href {https://doi.org/10.1109/TKDE.2020.2981314} {A survey on deep learning for named entity recognition}.
\newblock \emph{IEEE Transactions on Knowledge and Data Engineering}, 34(1):50--70.

\bibitem[{Lourentzou et~al.(2019)Lourentzou, Manghnani, and Zhai}]{lourentzou2019adapting}
Ioannis Lourentzou, Kiran Manghnani, and ChengXiang Zhai. 2019.
\newblock Adapting sequence to sequence models for text normalization in social media.
\newblock In \emph{Proceedings of the international AAAI conference on web and social media}, volume~13, pages 335--345.

\bibitem[{Mao and Liu(2019)}]{mao2019bert}
Jie Mao and Wenbo Liu. 2019.
\newblock A bert-based approach for automatic humor detection and scoring.
\newblock In \emph{IberLEF@SEPLN}, pages 197--202.

\bibitem[{Pal and Saha(2015)}]{pal2015detection}
Anirban~Roy Pal and Debajyoti Saha. 2015.
\newblock Detection of slang words in e-data using semi-supervised learning.
\newblock \emph{arXiv preprint arXiv:1702.04241}.

\bibitem[{Pei et~al.(2019)Pei, Sun, and Xu}]{pei2019slang}
Zhen Pei, Zhiyuan Sun, and Yang Xu. 2019.
\newblock Slang detection and identification.
\newblock \emph{Proceedings of the 23rd Conference on Computational Natural Language Learning (CoNLL)}, pages 881--889.

\bibitem[{Wu et~al.(2018)Wu, Morstatter, and Liu}]{wu2018slangsd}
Lingfei Wu, Fred Morstatter, and Huan Liu. 2018.
\newblock Slangsd: building, expanding and using a sentiment dictionary of slang words for short-text sentiment classification.
\newblock \emph{Language Resources and Evaluation}, 52:839--852.

\bibitem[{Zeng et~al.(2022)Zeng, Liu, Du, Wang, Lai, Ding, Yang, Xu, Zheng, Xia et~al.}]{zeng2022glm}
Aohan Zeng, Xiao Liu, Zhengxiao Du, Zihan Wang, Hanyu Lai, Ming Ding, Zhuoyi Yang, Yifan Xu, Wendi Zheng, Xiao Xia, et~al. 2022.
\newblock Glm-130b: An open bilingual pre-trained model.
\newblock \emph{arXiv preprint arXiv:2210.02414}.

\bibitem[{Zhang et~al.(2015)Zhang, Zhao, and LeCun}]{zhang2015character}
Xiang Zhang, Junbo Zhao, and Yann LeCun. 2015.
\newblock Character-level convolutional networks for text classification.
\newblock \emph{Advances in Neural Information Processing Systems}, 28:649--657.

\end{thebibliography}
\bibliographystyle{acl_natbib}

\end{document}